\documentclass[conference]{IEEEtran}
\usepackage{cite}
\usepackage{amsmath,amssymb,amsfonts}
\usepackage{algorithmic}
 \usepackage{hyperref}
\usepackage{graphicx}
\usepackage{textcomp}
\usepackage{booktabs}
\usepackage{xcolor}
\def\BibTeX{{\rm B\kern-.05em{\sc i\kern-.025em b}\kern-.08em
    T\kern-.1667em\lower.7ex\hbox{E}\kern-.125emX}}
\begin{document}

\title{Personalized Automatic Speech Recognition \\ for a Dysarthric and Tracheostomic Speaker \\ using Artificial Conversations
}

\author{
\IEEEauthorblockN{David Nadrchal, Monorama Swain, Florian Schmid, Gerhard Widmer and Paul Primus}
\IEEEauthorblockA{\textit{Institute of Computational Perception (CP-JKU)} \\
\textit{Johannes Kepler University Linz}\\
Linz, Austria \\
ORCIDs: 0009-0000-3208-3355 (D.N.), 0009-0000-4663-6141 (M.S.), 0009-0007-4988-6067 (F.S.), \\
0000-0003-3531-1282 (G.W.), 0009-0000-9889-8493 (P.P.)}
}

\IEEEoverridecommandlockouts
\IEEEpubid{\makebox[\columnwidth]{\parbox{\columnwidth}{\vspace{10pt}\copyright~2026 IEEE. Personal use is permitted, but republication/redistribution requires IEEE permission.}\hfill} \hspace{\columnsep}\makebox[\columnwidth]{ }}

\maketitle

\begin{abstract}
This work presents an automatic speech recognition (ASR) system personalized for a Czech speaker with a permanent tracheal stoma and severe dysarthria rendering their speech unintelligible to untrained listeners. We release a public dataset containing 33 annotated hours of the speaker's speech, collected using a novel "artificial conversation" protocol designed for high engagement and dialogue realism. We propose a multi-stage training pipeline based on Whisper Base: fine-tuning on standard Czech speech, acoustically simulated tracheostomic speech, and the speaker's data. We evaluate the system across three near real-time scenarios: scripted conversations, question answering, and spontaneous dialogue, achieving a 50\% relative reduction in Character Error Rate compared to Whisper Base baseline and surpassing the average recognition accuracy of their assistants in acoustic recognition of isolated utterances. We demonstrate that even for severely impeded speech, a helpful ASR is achievable, as evidenced by the quantitative results and the feedback from the speaker.
\end{abstract}


\section{Introduction}

\begin{figure}[b!]
    \centering
    \includegraphics[width=0.85\linewidth]{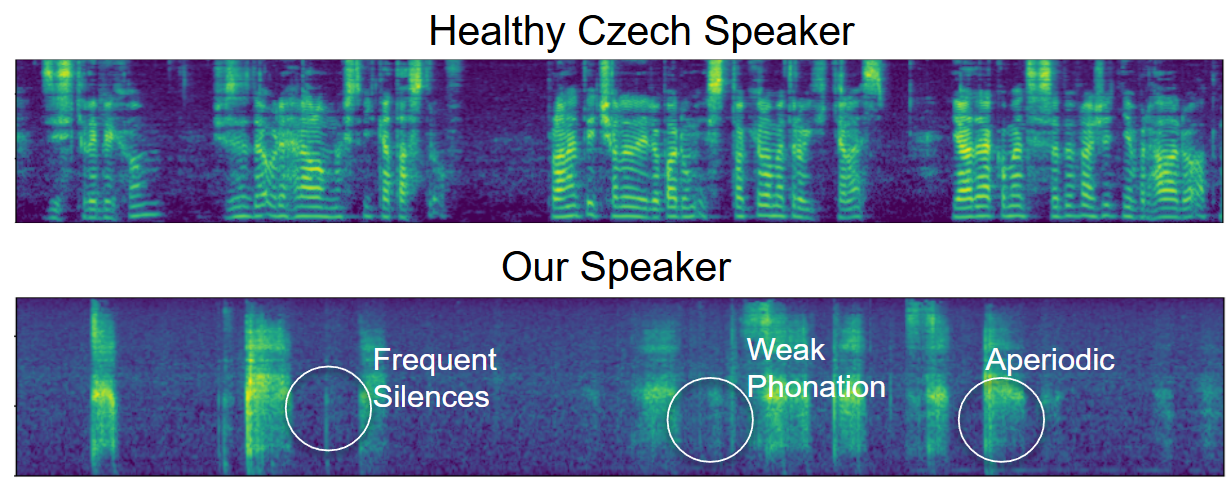}
    \caption{Speech disorders of the target speaker - Mel spectrograms (128 bins) comparing their speech to that of a standard speaker.}
    \label{fig:speech_scheme}
\end{figure}

The speaker is a Czech teenager with quadriplegia and severe dysarthria following a spinal injury. They require a permanent tracheal stoma, rendering their speech unintelligible to untrained listeners (as quantified in Section~\ref{ss:human_intelligibility_assesment}). However, family members and experienced caregivers can understand the speaker in real-world interactions where non-verbal communication, lip-reading, and feedback loops are employed. We illustrate the spectral characteristics of the speaker's speech—such as reduced energy, spectral flatness, and frequent interruptions—in Fig.~\ref{fig:speech_scheme}. These  physical limitations prevent the speaker from fully participating in education alongside their peers, as existing assistive technologies do not adequately address their communication needs.

Low-tech Augmentative and Alternative Communication (AAC) methods, such as blink codes or communication boards~\cite{millar_1994}, are too slow for engaging, real-time interaction; indeed, the speaker can already communicate faster by nose-typing on their smartphone.

More recently, assistive Automatic Speech Recognition (ASR) systems have emerged as powerful high-tech AAC methods. Pioneering initiatives like Project Euphonia~\cite{macdonald_2021} and Voiceitt~\cite{howarth_2024} demonstrated the viability of personalized ASR. Building on this, the 2025 Speech Accessibility Project~\cite{hasegawa_2024} (SAP) challenge recently saw the top-performing team~\cite{takahashi_2025} achieve an 8\% Word Error Rate (WER) on impeded speech in unconstrained ASR settings using fine-tuned Prakeet-TDT model~\cite{sekoyan_2025}. Unfortunately, these solutions remain largely inapplicable to our case, as they focus heavily on English and rely on large-scale data collection across multiple speakers. Our scenario requires a highly personalized approach for a single speaker with severe, combined dysarthria and tracheostomic phonation, for which no prior Czech training data exist.

To quantify the difficulty of this specific acoustic modeling task, we evaluated the speaker's baseline intelligibility using Whisper large-v3~\cite{radford_2023}, a state-of-the-art multilingual ASR foundation model. The model failed to recognize a single word correctly, yielding a WER exceeding 100\% due to rampant insertion errors. For context, the same off-the-shelf Whisper model achieves a 23.64\% WER on the English SAP dataset. We provide a more rigorous intelligibility assessment in Section~\ref{s:results}. 

Furthermore, while recent advances in speech neuroprostheses (e.g., Card et al.~\cite{card_2024}) have achieved a remarkable 2.5\% WER for ALS patients using neural implants, such methods are expensive and highly invasive, rendering them impractical for our user's immediate real-world needs.

To address these challenges, we propose a personalized, non-invasive ASR system developed through a multi-stage pipeline. Our main contributions are:

\begin{enumerate}
\item \textbf{Public Dataset:} We release 33 hours of annotated, severely dysarthric and tracheostomic Czech speech. To the best of our knowledge, this is the largest publicly available single-speaker corpus for impeded speech.
\item \textbf{Artificial Conversations:} We introduce a novel data collection protocol that simulates natural dialogue dynamics. This method maintains speaker engagement and elicits natural prosody while retaining the tractability of scripted targets, enabling highly efficient data collection by a single researcher (Section~\ref{s:dataset}).
\item \textbf{Real-Time Streaming Framework:} Unlike prior work focused primarily on offline assistive ASR, we propose a near-real-time inference pipeline. This includes a device-adaptation fine-tuning stage, a personalized Voice Activity Detector (VAD) optimized for very low signal-to-noise ratios, and guided language decoding to enable practical, real-world smartphone deployment.
\end{enumerate}

We evaluate the system across three live recognition scenarios: scripted conversations, question answering, and spontaneous dialogue (Section~\ref{s:experiments}). We achieve performance on par with experienced listeners in terms of recognition of isolated audio-only utterances, concluding that for further reduction of the error we must consider also other context than the spoken utterances themselves such as longer conversation history or visual input. Nevertheless, the resulting system is considered useful by the speaker. 

We provide the codes for dataset collection, the machine learning pipeline and the web interface for streaming, near-real time inference on our GitHub~\footnote{ https://github.com/Hobit2002/DysartrTracheoASR}. 

\section{Dataset}
\label{s:dataset}

\subsection{Artificial Conversations}

Collection of speech data often relies on the participants reading provided sentences~\cite{ardila_2020, macdonald_2021, panayotov_2015}. It has been shown that such speech substantially differs from that of our target domain - real-world conversations~\cite{nakamura_2008}. Furthermore, the speaker reported that standard reading sessions lacked the engagement necessary to sustain long-term data collection (to quote them: "\textit{It's a drag. I constantly feel such powerlessness and a sense of futility.}"). To address this, we introduce a novel collection protocol: Artificial Conversations.

\begin{figure}
    \centering
    \includegraphics[width=0.85\linewidth]{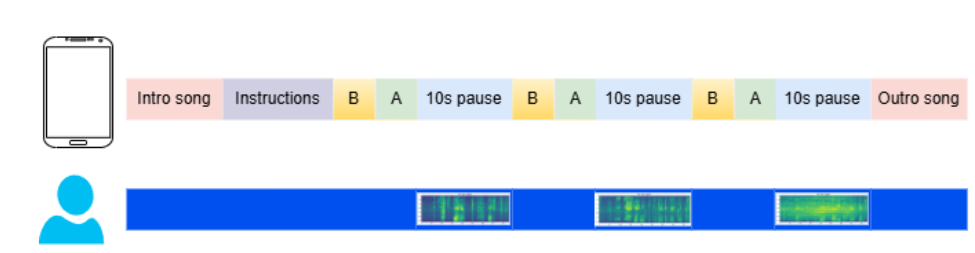}
    \caption{Artificial conversations - the upper track depicts the synthesized audio guide provided to the speaker. The lower track illustrates the silence windows where the speaker was expected to repeat their line of the scripted dialogue.}
    \label{fig:artificial_conversations_scheme}
\end{figure}

This method simulates natural dialogue dynamics while retaining the benefits of scripted targets. We provide the speaker with a pre-rendered audio track containing a dialogue between two characters (A and B). The track includes synthesized speech for Character B and silence intervals for Character A. The speaker's task is to "play" Character A by repeating the target line during the allocated silence (see Fig.~\ref{fig:artificial_conversations_scheme}).

Because existing corpora are not sufficiently personalised to keep the speaker engaged and to contain the vocabulary tailored to their communication needs, we generate the scripts with a language model to cover the topics of their interest. We prompt GPT-4o mini~\cite{openai_2024} (chosen for a favorable tradeoff between speed and realism of generated dialogues) with six opening lines of the dialogue, 1--4 sentences describing its high-level topics and detailed instructions on the manner (tone, sentence length, vocabulary, etc.)  in which the characters should speak. The resulting dialogue is split into lines spoken by character A and lines spoken by character B. Both sets of lines are converted to audio using a text-to-speech system. The lines spoken by character B are further processed with ring modulation to make them distinguishable from those of character A.  Finally, we merge the audio into a single dialogue file. While creating the audio file, we also save the temporal coordinates of the silence segments during which the speaker is expected to speak, along with the corresponding lines of text. These timestamps are later used to annotate the recording.

In order to make the "artificial conversations" both engaging and representative of the speaker's real-world vocabulary, we dedicate most of the recording sessions to topics revolving around the speaker's interests and everyday life.

\subsection{Dataset Statistics}

We conducted 137 recording sessions and collected 16,619 sentence-level utterances totalling 37:45 hours of transcribed speech. The sessions were predominantly "artificial conversations" but we have also recorded several reading sessions, manually transcribed spontaneous conversations, open-ended question answering (QA) or singing. The data distribution by recording modality is detailed in Table~\ref{tab:data_stats}. 

A subset of the "artificial conversations", spontaneous conversations and QA sessions were recorded in streaming settings, where the speaker interacted directly with our real-time ASR prototype. We refer to those sessions and the speech data resulting from them as "streaming". The remainder were collected in an offline setting (processed post-hoc). "Artificial conversations" were annotated automatically using the timestamp files created during the generation process; other formats were transcribed manually. 

\begin{table}[h]
    \centering 
    \caption{Overall Dataset Statistics} 
    \label{tab:data_stats} 
    \begin{tabular}{lccc}
        
        \textbf{Session Type} & \textbf{Sessions} & \textbf{Hours} & \textbf{Utts} \\
        
        Artif. Convs. (Offline) & 69 & 20:21 & 8 428 \\
        Artif. Convs. (Streaming) & 30 & 10:39 & 2 733   \\
        Reading & 21 & 3:06 & 2 719 \\
        Question Answering & 3 & 0:48 & 353  \\
        Spontaneous Convs. & 4 & 1:35 & 737  \\
        Other (Singing, etc.) & 10 & 1:16 & 454 \\
        
        \textbf{Total} & \textbf{137} & \textbf{37:45} & \textbf{16 619} \\ 
         
    \end{tabular}
\end{table}

Although reading sessions yielded slightly more utterances per session than artificial conversations (129.5 vs.\ 112.7), artificial conversations produced nearly double the word count (420.1 vs.\ 229.9 words per session). Given our collection rate of approximately three sessions per week for both formats, artificial conversations proved significantly more data-efficient for our particular case.

We release a subset of this data on Zenodo\footnote{https://zenodo.org/records/22745518}, excluding personally identifiable information, with explicit consent from the speaker and their legal guardian. Each utterance includes session identifiers and dates to enable longitudinal analysis and prevent data leakage.

For evaluation, we used a session-wise split to avoid channel or condition leakage, reserving held-out streaming sessions for testing to reflect real-world deployment. The remaining sessions form the training and validation sets; the test data distribution is detailed in Table~\ref{tab:test_stats}. All of the test data were recorded in the streaming settings.

\begin{table}[h]
    \centering 
    \caption{Test Set Statistics (Subset of the Overall Dataset)} 
    \label{tab:test_stats} 
    \begin{tabular}{lccc}
        
        \textbf{Session Type} & \textbf{Sessions} & \textbf{Hours} & \textbf{Utts} \\
        
        Artif. Convs  & 3 & 1:08 & 329   \\
        Question Answering & 3 & 0:48 & 353  \\
        Spontaneous Convs. & 1 & 0:23 & 188  \\
         
    \end{tabular}
\end{table}

\section{Method}
\label{s:method}

\subsection{Training Pipeline}
\label{ss:training_pipeline}

We build our system upon the off-the-shelf Whisper Base provided by OpenAI~\cite{radford_2023}. While it generally yields lower accuracy than larger models like Whisper Large~\cite{radford_2023} or Parakeet~\cite{sekoyan_2025}, we selected it for its multilingual capabilities and low computational overhead, which enables near real-time inference even with limited hardware resources. We carry out the following four-stage pipeline to fine-tune this pretrained model on regular, synthesized~\cite{violeta_2024} and target speaker's speech data:

\begin{enumerate}
    \item \textbf{Standard Czech Speech}: To enhance its understanding of Czech language, we fine-tune the off-the-shelf Whisper Base on the Czech split of Common Voice 19.0~\cite{ardila_2020} and on the Ortofon-v3~\cite{lukes_2024}, a dataset of 243 hours of informal spoken Czech speech.
    \item \textbf{Simulated Tracheostomic Speech}: To compensate for limited data for our speaker, we simulate their condition on standard Czech speech. We provide more detail on the simulation in Subsection~\ref{ss:acoustic_simulation}.
    \item \textbf{Offline Collected Speaker's Data}: We fine-tune the model on 21:33 hours of the offline collected speech of the speaker.
    \item \textbf{Device Adaptation on Streaming Data}: Final tuning on 5:49 hours of streaming data to adapt to the specific noise suppression and microphone characteristics of the deployment device.
\end{enumerate}

We use the Adam optimizer across all stages, applying a learning rate of $3 \cdot 10^{-5}$ for the first three stages and reducing it to $1 \cdot 10^{-5}$ for the final streaming adaptation (complete hyperparameters are available in the linked repository). To prevent linguistic degradation, we regularize training in all stages using Knowledge Distillation from a language model (detailed in Section~\ref{ss:kd_from_lm}). Further methodologies for acoustic simulation and streaming adjustments are discussed in Sections~\ref{ss:acoustic_simulation} and~\ref{ss:streaming_interface}, respectively.

\subsection{Knowledge Distillation from a Language Model} \label{ss:kd_from_lm}
\begin{figure}[h]
    \centering
    \includegraphics[width=0.8\linewidth]{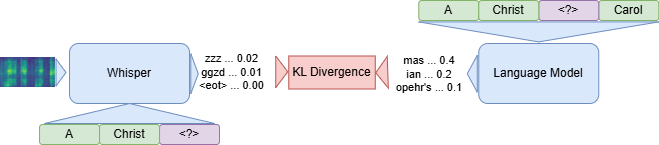}
    \caption{KD mechanism.}
    \label{fig:mlm_kd_scheme}
\end{figure}

To counteract Whisper's poor Czech prior, we regularize fine-tuning via Knowledge Distillation~\cite{hinton_2015, futami_2020} from a custom 60M-parameter, 6-layer bi-LSTM language model (LM). This LM shares Whisper's BPE tokenizer~\cite{gage_1994} and predicts masked tokens using bidirectional context~\cite{devlin_2019}. During fine-tuning, the LM's target probability distributions are distilled into Whisper via KL divergence (Fig.~\ref{fig:mlm_kd_scheme}; hyperparameters on GitHub).

\subsection{Acoustic Simulation}
\label{ss:acoustic_simulation}

To overcome the scarcity of tracheostomic speech data, we augment non-impaired speech from Common Voice and Ortofon with acoustic transformations that mimic the condition of the target speaker. The standard deep learning voice conversion methods (zero-shot Seed-VC~\cite{liu_2024}, non-parallel CycleGAN-VC2~\cite{kaneko_2019}, and self-supervised S3PRL-VC~\cite{huang_2022}) failed to adequately model this extreme phonetic degradation. We attribute this to severe domain mismatch: tracheostomic aphonia falls far outside the distribution of standard pre-trained latent extractors. Consequently, we opted for deterministic, low-level signal processing transformations. This choice is further supported by recent literature~\cite{piernicki_2026}, which suggests that ASR pre-training on synthetic data generated by complex neural techniques is susceptible to overfitting in downstream tasks.

We parameterize these transformations using a local hill-climbing search~\cite{russel_2010} maximizing the acoustic indistinguishability between simulated and real samples. We heuristically quantify this by K-means clustering their embeddings (PyAnnote~\cite{bredin_2020} canonical x-vectors~\cite{snyder_2018}) to two clusters and maximizing the mean intra-cluster Shannon entropy~\cite{shannon_1948}, aiming to transition from class-separated clusters (entropy $\approx$ 0) to a uniform mix (entropy $\approx$ 1). We consider this metric to be only a coarse proxy that might not reflect the acoustic similarity (especially for the out-of-distribution speech such as that of our speaker). The simulation consists of three steps:

\begin{enumerate}
    \item \textbf{Spectral Flattening:} We isolate the periodic voice component using Linear Predictive Coding~\cite{oshaughnessy_1988} and replace it with white noise to simulate aphonia.
    \item \textbf{Temporal Alignment:} We slow the audio by a factor of $4\times$ using librosa phase vocoder time stretch to match the speaker's rate of speech ($\approx$25--30 wpm).
    \item \textbf{Interruption Simulation:} We inject random 250\,ms silence segments to mimic dysfluent pauses of the speaker.
\end{enumerate}

On our experimental set this pipeline increased the embedding cluster entropy from 0.5 to 0.7--0.8. Listening tests also indicated a shift towards tracheostomic speech, although the differences remained notable.

We also experiment with a simpler acoustic simulation that only uses the time-stretching, as it was shown that this augmentation alone can lead to significant performance gains~\cite{prananta_2022}. The comparison to the full acoustic simulation pipeline is provided in Section~\ref{s:results}

\subsection{Streaming Inference}
\label{ss:streaming_interface}

To deploy the system for real-time use, we developed a web-based interface accessible via smartphone. Audio is streamed to our inference server in 62.5\,ms chunks (1000 samples at 16 kHz) for continuous processing.

While the latency of our system ($\approx$2.5\,s) is higher than the sub-second commercial standard~\cite{sainath_2020}, the user reported that it remains acceptable, as it is faster than the time typically required for human listeners to decode their speech.

The exact settings of detection thresholds and other inference hyperparamters listed below were experimentally determined by CER minimisation on a validation set.

\subsubsection{Personalized Voice Activity Detection}

To enable hands-free interaction, we find it necessary for our system to employ VAD. As pretrained solutions such as PyAnnote 2.1.~\cite{bredin_2021} fail to reliably distinguish the speaker's weak phonation from non-speech sounds~\cite{awan_2009}, we implement a personalized VAD classifier using Logistic Regression, trained on features extracted from two of the streaming "artificial conversations". We use 1057 of one second manually segmented and annotated training chunks, 533 of which correspond to "silence" and 524 to "speech" that are split into 62.5ms chunks. For each audio chunk (zero-padded to 2048 samples), we extract spectral flatness, zero-crossing rate, spectral centroid, and 13-bin MFCCs. The classifier uses a detection threshold of 0.3. Positive detections are grouped into utterance segments, allowing for intra-utterance pauses of up to 2.0\,s. Segments exceeding 1.5\,s are padded with 1.0\,s of buffered preceding context and 2.0\,s of trailing context (accumulated during the silence timeout) before causal inference. 

On the validation set, these measures largely closed the gap in CER between automatic and manual speech segmentation. The test set evaluation is provided in Section~\ref{s:results}. Thereby, we also provide a comparison to Pyannote VAD fine-tuned on our single-session training set.

\subsubsection{Decoding and Constraints}

We employ a beam search with a width of 25 to maximize transcription accuracy despite the acoustic ambiguity. To mitigate phonetic hallucinations, we constrain decoding to a standard Czech lexicon~\cite{altun_2023} augmented with the vocabulary of training sessions using Trie-based logit biasing~\cite{zhao_2019}. We penalize the probabilities of out-of-vocabulary tokens by a factor of $10^{-4}$, guiding the model towards expected vocabulary.

Finally, to filter out residual non-speech noise, we discard transcriptions with an average token probability lower than the experimentally determined threshold of 0.712. 

\section{Experimental Setup}
\label{s:experiments}

To evaluate our system, we define three test scenarios ranging from scripted to spontaneous speech. For each scenario, we establish a parallel non-impaired speech baseline using an off-the-shelf Whisper Base model. This allows us to disentangle errors caused by the speech impairments from those caused by the difficulty of the task itself. We compare these results with the intelligibility of the speaker's speech to human Czech listeners.

\subsection{Evaluation Protocol}

To obtain an estimate of system performance reflecting challenges related to real-world setup, the pathological setup includes automatic VAD and segmentation on continuous streaming audio. In contrast, the non-impaired speech baseline are evaluated on manually segmented utterances. This avoids inflating performance through oracle segmentation for the target speaker.

We report performance using Character Error Rate (CER), calculated as:

\begin{equation}
\text{CER} = \frac{S + D + I}{N}, \label{eq:cer}
\end{equation}

where $S$, $D$, and $I$ are the number of character substitutions, deletions, and insertions, respectively, and $N$ is the total number of characters in the reference transcription. Note that the model may hallucinate text, causing insertions ($I$) to exceed the reference length ($N$). Thus \eqref{eq:cer} can yield values greater than 1.0. We prefer CER over Word Error Rate (WER) for two reasons: (1) Czech is a highly inflected language where minor suffix errors heavily penalize WER, and (2) in high-noise pathological scenarios, CER provides more granular insight into phonetic preservation than binary word-level hits~\cite{thennal_2025}.

\subsection{Test Scenarios}

\subsubsection{Scripted Conversations}

We utilize 3 streaming artificial conversations (329 utterances). To create a non-impaired speech reference, we re-recorded these exact scripts under similar acoustic conditions. This results in a pair of datasets with identical linguistic content but differing vocal characteristics.

\subsubsection{Question Answering (QA)}

We evaluate on 3 QA sessions consisting of 353 short utterances (often single-word answers). Ground truth transcriptions were generated by an expert human listener. As with the scripted scenario, we created a parallel non-impaired baseline by re-recording the answers to ensure identical content. 

\subsubsection{Spontaneous Conversations}

This represents the most challenging "in-the-wild" benchmark. 

\begin{itemize}
    \item Pathological Set: We use one fully transcribed spontaneous conversation (188 utterances) recorded in the streaming settings. To isolate the patient's speech from their interlocutor, we apply speaker diarization using PyAnnote~\cite{bredin_2020} prior to our VAD pipeline.
    \item Non-impaired Baseline: We find it is unfeasible to replicate a spontaneous conversation while keeping its spontaneity. Therefore, instead of strict re-recording as above, we use a proxy baseline: 16 recordings (5 136 utterances) from the Ortofon-v3 dataset~\cite{lukes_2024}, representing natural, informal Czech conversations. These were withheld from the traning set of stage \textit{(1)}.
\end{itemize}

\section{Results and Analysis}
\label{s:results}

We evaluate our system by benchmarking performance against human listeners (Section~\ref{ss:human_intelligibility_assesment}) and standard Czech Whisper baselines (Section~\ref{ss:comparison_to_baseline}). We then assess individual system components (Section~\ref{ss:componentwise_evaluation}) and present a qualitative error analysis (Section~\ref{ss:qualitative_error_analysis}).

\subsection{Human Intelligibility Assessment}
\label{ss:human_intelligibility_assesment}

To contextualize our system's performance, we evaluated the baseline intelligibility of the speaker's speech to human listeners. We conducted a listening experiment in which eight Czech speakers, unfamiliar with the target speaker, transcribed 25 utterances. These utterances consisted of isolated one- to three-word recordings, randomly sampled from three distinct streaming artificial conversations and presented in a shuffled order. The untrained listeners yielded an average CER of 1.182 (exceeding 1.0 due to insertion errors), confirming that the speaker's speech is effectively unintelligible to naive listeners in an audio-only setting.

To establish a baseline for experienced listeners, we further recruited four of the speaker's assistants. They achieved an average CER of 0.708 (range 0.42--0.90). A Wilcoxon signed-rank test~\cite{wilcoxon_1992} confirms that our system (which achieved a CER of 0.565 on this specific utterance set) significantly outperforms both the group average ($p=0.046$) and three of the individual assistants ($p < 0.01$). However, the top-performing human listener (CER 0.423) still significantly surpasses the model ($p=0.008$), indicating that a performance gap remains compared to ultimate expert-level comprehension. 

Notably, this top-performing listener was one of the authors who curated the dataset. Consequently, they possessed strong prior knowledge of the expected linguistic content and, unlike the others assistants, were accustomed to decoding the speaker's audio-only speech without relying on visual cues or an interactive feedback loop. 

\subsection{Comparison to Baseline}
\label{ss:comparison_to_baseline}

Table~\ref{tab:main_results} compares our system against standard Czech models (Baseline, Non-impaired Topline) and a "Speaker-only" baseline. This baseline represents a version of our system fine-tuned exclusively on the speaker's data, omitting the initial pre-training stages (skipping to steps 3 and 4 of the pipeline described in Section~\ref{ss:training_pipeline}).

While our proposed system reduces the Character Error Rate (CER) by 40 to 90 absolute percentage points compared to the zero-shot baseline, a gap remains when compared to non-impaired speech recognition. Furthermore, while omitting the pre-training (i.e., the "Speaker-only" configuration) does not substantially impact performance for scripted conversations, it severely degrades accuracy across the less constrained modalities. We suggest that this ambiguous effect might be because the pre-training steps establish language priors for spontaneous speech that are effective for spontaneous settings but not for the scripted conversations that likely follows very similar acoustic and linguistic patterns as the training conversations.

\begin{table}[b!]
    \centering
    \caption{Comparison of CER across test scenarios. "Baseline" is off-the-shelf Whisper Base. "Non-impaired Topline" represents the theoretical upper bound performance: the Whisper Base model fine-tuned by us on standard Czech speech (Common Voice and Ortofon). "Speaker-only" stands for a model fine-tuned only on the speaker's speech without the pre-training stages.}
    \label{tab:main_results}
    \resizebox{\linewidth}{!}{%
    \begin{tabular}{lccc}
        
        \textbf{Evaluation Setup} & \textbf{Scripted} & \textbf{QA} & \textbf{Spont.} \\ 
        
        \multicolumn{4}{l}{\textit{Non-impaired Control Data}} \\ 
        \quad Baseline  & 0.289 & 0.286 & 0.710 \\  
        \quad Non-impaired Topline (Czech FT) & 0.197 & 0.172 & 0.256 \\ 
        
        \multicolumn{4}{l}{\textit{Target Speaker Data (Impeded)}} \\ 
        \quad Baseline (Zero-shot) & 0.944 & 0.882 & 1.529 \\
        \quad Speaker-only & \textbf{0.431} & 0.510 & 0.949 \\
        \quad \textbf{Proposed System (Ours)} & 0.433 & \textbf{0.449} & \textbf{0.637} \\ 
        
    \end{tabular}%
    }
\end{table}

\subsection{Component-wise Evaluation}
\label{ss:componentwise_evaluation}

Table~\ref{tab:ablation} details the contribution of specific system components. \textit{Device Adaptation} and \textit{Inference Components} (Personalized VAD, Vocab Guidance) proved critical; removing either degrades performance by $>27\%$ (relative). We attribute the significance of \textit{Device Adaptation} to the aggressive noise suppression filters employed by the iPhone in streaming settings.

Our lightweight VAD outperforms \textit{Ft. Pyannote VAD} (Pyannote VAD fine-tuned on one of the streaming sessions) across all three modalities, possibly because the PyanNet network~\cite{bredin_2020} overfitted the limited training data. The \textit{Manual VAD Oracle} (replacing our custom VAD with handcrafted speech segmentations unconstrained by causal streaming requirements) reveals a noticeable performance gap in spontaneous speech (0.506 vs. 0.637 CER). This suggests that $\approx$20\% of the error in this scenario stems purely from segmentation challenges. However, qualitative analysis reveals that this increased error rate is largely driven by false positives—specifically, the simple VAD classifier incorrectly triggering on the speech of the interlocutor. From a usability perspective, these false triggers do not severely disrupt the actual conversation flow, as they occur while the communication partner is speaking—a period when neither participant is typically monitoring the smartphone's transcription screen. Therefore, while there remains an algorithmic gap in dynamic speaker diarization, our lightweight VAD implementation does not constitute a practical bottleneck for real-world deployment. 

\textit{Acoustic Simulation} and \textit{Pre-training} provided moderate gains (3--6\%), while \textit{Knowledge Distillation} offered mixed results across formats, especially considering the very limited number of our test samples. Notably, when substituting our full simulation for a \textit{Stretch-only Acoustic Simulation}, performance drops drastically in spontaneous settings, despite minor improvements in the other modalities. To explain this discrepancy, we hypothesize that during controlled recordings eliciting short answers, the speaker maintains largely vocalized speech. In contrast, during real, spontaneous conversations, the speaker frequently transitions to whispered, aperiodic speech---a condition that directly benefits from the removal of the periodic source during our full acoustic simulation phase.

Finally, Table~\ref{tab:progressive_ablation} illustrates the performance trajectory across our multi-stage pipeline. While error rates for scripted speech decrease steadily throughout the pipeline, QA and spontaneous speech are only recognized with reliable accuracy after training the model on the target speaker's offline data.

\begin{table}[t]
    \centering
    \caption{Ablation study. Results reported in CER. The \textbf{Avg. Rel. $\Delta$} column represents the average percentage increase in error relative to the Full System (higher implies the component is more critical). The best result in each column is \textbf{bolded}, and the second best is \underline{underlined}.}
    \label{tab:ablation}
    \resizebox{\linewidth}{!}{%
    \begin{tabular}{lcccc|c}
        
        \textbf{Configuration} & \textbf{Scripted} & \textbf{QA} & \textbf{Spont.} & \textbf{Mean} & \textbf{Avg. Rel. $\Delta$} \\ 
        \textbf{Full System} & 0.433 & \underline{0.449} & \underline{0.637} & \textbf{0.506} & \textit{Reference} \\ 
        
        \textit{Training Components} & & & & & \\
        \quad w/o Device Adaptation & 0.637 & 0.637 & 0.696 & 0.657 & +32.7\% \\
        \quad w/ Stretch-only Ac. Simulation & \textbf{0.411} & \textbf{0.437} & 0.869 & 0.572 & +9.6\% \\
        \quad w/o Acoustic Simulation & 0.453 & 0.500 & 0.647 & 0.533 & +5.8\% \\
        \quad w/o Common Voice Pre-train & 0.471 & 0.475 & 0.645 & 0.530 & +5.2\% \\
        \quad w/o Ortofon Pre-train & 0.438 & 0.451 & 0.695 & 0.528 & +3.5\% \\
        \quad w/o Knowledge Distillation & \underline{0.428} & 0.499 & \textbf{0.634} & \underline{0.520} & +3.2\% \\
        
        \textit{Inference Components} & & & & & \\
        \quad w/o Personalized VAD & 0.574 & 0.588 & 0.793 & 0.652 & +29.3\% \\ 
        \quad w/ Ft. Pyannote VAD & 0.599 & 0.594 & 0.659 & 0.617 & +24.7\% \\
        \quad w/o Vocab. Guidance & 0.572 & 0.583 & 0.763 & 0.639 & +27.2\% \\
        
        \textit{w/ Manual VAD Oracle} & \textit{0.409} & \textit{0.416} & \textit{0.506} & \textit{0.444} & \textit{--} \\
        
    \end{tabular}
    }
\end{table}

\begin{table}[t]
    \centering
    \caption{Incremental evaluation across the multi-stage training pipeline. Results are reported in Character Error Rate (CER).}
    \label{tab:progressive_ablation}
    \resizebox{\linewidth}{!}{%
    \begin{tabular}{lcccc}
        
        \textbf{Training Stage (Cumulative)} & \textbf{Scripted} & \textbf{QA} & \textbf{Spont.} & \textbf{Mean} \\ 
        Whisper Base (Zero-shot) & 0.944 & 0.882 & 1.529 & 1.118 \\
        \quad + Czech FT (Standard Speech) & 0.904 & 1.796\rlap{\textsuperscript{*}} & 2.633\rlap{\textsuperscript{*}} & 1.778\rlap{\textsuperscript{*}} \\
        \quad + Simulated Tracheostomic Speech & 0.772 & 1.082\rlap{\textsuperscript{*}} & 5.358\rlap{\textsuperscript{*}} & 2.404\rlap{\textsuperscript{*}} \\
        \quad + Offline Speaker Data & 0.637 & 0.637 & 0.696 & 0.657 \\
        \quad + Streaming Data (Full System) & \textbf{0.433} & \textbf{0.449} & \textbf{0.637} & \textbf{0.506} \\ 
        
        \multicolumn{5}{l}{\scriptsize \textsuperscript{*}Values inflated heavily by unconstrained insertion errors (hallucinations) prior to speaker-specific grounding.}
    \end{tabular}
    }
\end{table}

\subsection{Qualitative Error Analysis}
\label{ss:qualitative_error_analysis}

Although the absolute CER remains high, semantic intent is often preserved as many errors stem missing diacritics or homophone substitutions (see Table~\ref{tab:qualitative_examples}). Applying aggressive text normalization to merge acoustically similar Czech characters (e.g., \textit{i}/\textit{y} or \textit{d}/\textit{t}) yields a 13.4\% relative CER reduction on scripted speech, dropping it to 0.375. As with non-impaired baselines, spontaneous speech remains the most challenging domain. 

Subjective feedback from the speaker highlights the practical utility: "\textit{It is better than it used to be; it really understands me... essentially most of the time. However, it falls short on the names of rockets and space shuttles—though the vast majority of people fall short on those too}."

\begin{table}[t]
    \centering
    \caption{Qualitative analysis of a sample transcript.  Errors are marked in \textcolor{red}{red}.}
    \label{tab:qualitative_examples}
    \resizebox{\linewidth}{!}{%
    \begin{tabular}{ll}
        
        \textbf{Reference (Ground Truth)} & \textbf{Hypothesis (System Output)} \\
        
        \textit{(Translation provided below)} & \\
        
        Ahoj, Šalino & Ahoj, Šalino \\
        \textit{\small{Hi, Šalina}} & \textit{\small{Hi, Šalina}} \\
        \addlinespace
        Lidé zapomínají na oblohu & Lidé zapomínají na oblohu \\
        \textit{\small{People forget about the sky}} & \textit{\small{People forget about the sky}} \\
        \addlinespace
        \textbf{Noční} oblohu & \textcolor{red}{\textbf{Naští}} oblohu \\
        \textit{\small{\textbf{Night} sky}} & \textit{\small{\textbf{[Nonsense]} sky}} \\
        \addlinespace
        Martin Gembec \textbf{doporučuje hvězdy} & Martin Gembec \textcolor{red}{\textbf{do položuje}} \\
        \textit{\small{Martin Gembec \textbf{recommends stars}}} & \textit{\small{Martin Gembec \textbf{[hallucinated verb phrase]}}} \\
        \addlinespace
        Tomáš Přibyl \textbf{má} skvělé \textbf{tipy} & Tomáš Přibyl \textcolor{red}{\textbf{na}} skvělé \textcolor{red}{\textbf{typy}} \\
        \textit{\small{Tomáš Přibyl \textbf{has} great \textbf{tips}}} & \textit{\small{Tomáš Přibyl \textbf{on} great \textbf{types}}} \\
        
    \end{tabular}
    }
\end{table}

\section{Conclusion}

We presented a personalized ASR system for a Czech speaker with severe dysarthria and a permanent tracheal stoma rendering their speech entirely unintelligible to both human listeners and state-of-the-art ASR models. To address the data scarcity of their condition, we introduced "Artificial Conversations"—a novel data collection protocol combining the engagement of dialogue with the tractability of scripted speech and released 33 hours of impeded speech collected in this manner. Our four-stage fine-tuning pipeline, augmented by acoustic simulation and extensive adaptations to streaming settings improved lightweight Whisper Base from being completely unusable to being perceived as helpful as confirmed by both experimental results and the user feedback.

While the absolute CER ($\approx$0.50) remains high for general standards, our system outperforms experienced human listeners in low-context scenarios. Furthermore, although the state-of-the-art ASR models for English dysarthric speech achieve lower error rates than our system, we emphasize that our case is specific because of the data scarcity of Czech language, high severity and rarity of the speaker's impediment and low-latency constraints.

Future work should focus on generalizing this approach to multiple speakers and reducing the reliance on annotated data through advanced pre-training or semi-supervised learning on untranscribed speech. Additionally, integrating multimodal context (e.g., lip-reading) and conversational memory could help bridge the remaining gap between acoustic recognition and full intelligibility. Finally, evaluating the system across a broader test set would provide deeper insights into the relative contributions of individual pipeline components.

\section{Acknowledgment}
We would like to thank Radek Skarnitzl, Tomáš Bořil, Lucie Benešová and Václav Cvrček from Charles University for their advice on the removal of periodic component during acoustic simulation and for familiarising us with the Ortofon dataset. Finally, we are very grateful to the speaker themselves and their family for their extraordinary patience during participation in this project.

\section{AI-Generated Content Disclosure}
Gemini Pro was used to polish the text and provide structural feedback during the process of writing.

\bibliographystyle{IEEEtran}
\bibliography{mybib}

\end{document}